\documentclass[10pt,twocolumn,letterpaper]{article}

\usepackage[pagenumbers]{iccv} 

\definecolor{iccvblue}{rgb}{0.21,0.49,0.74}
\usepackage[pagebackref,breaklinks,colorlinks,allcolors=iccvblue]{hyperref}
\usepackage[accsupp]{axessibility}
\makeatletter
\def\blfootnote{\xdef\@thefnmark{}\@footnotetext}
\makeatother

\usepackage{bm}
\usepackage{graphicx}
\usepackage{subcaption}

\def\confName{ICCV}
\def\confYear{2025}

\title{Multimodal Prompt Alignment for Facial Expression Recognition}

\author{Fuyan Ma$^{1}$, ~Yiran He$^{2}$, ~Bin Sun$^3$, ~Shutao Li$^3$ \\
$^1$Chinese Academy of Military Science\\
$^2$Changchun University of Science and Technology, $^3$Hunan University\\
}

\begin{document}
\maketitle

\begin{abstract}
Prompt learning has been widely adopted to efficiently adapt vision-language models (VLMs) like CLIP for various downstream tasks. Despite their success, current VLM-based facial expression recognition (FER) methods struggle to capture fine-grained textual-visual relationships, which are essential for distinguishing subtle differences between facial expressions. To address this challenge, we propose a multimodal prompt alignment framework for FER, called MPA-FER, that provides fine-grained semantic guidance to the learning process of prompted visual features, resulting in more precise and interpretable representations. Specifically, we introduce a multi-granularity hard prompt generation strategy that utilizes a large language model (LLM) like ChatGPT to generate detailed descriptions for each facial expression. The LLM-based external knowledge is injected into the soft prompts by minimizing the feature discrepancy between the soft prompts and the hard prompts. To preserve the generalization abilities of the pretrained CLIP model, our approach incorporates prototype-guided visual feature alignment, ensuring that the prompted visual features from the frozen image encoder align closely with class-specific prototypes. Additionally, we propose a cross-modal global-local alignment module that focuses on expression-relevant facial features, further improving the alignment between textual and visual features. Extensive experiments demonstrate our framework outperforms state-of-the-art methods on three FER benchmark datasets, while retaining the benefits of the pretrained model and minimizing computational costs.
\end{abstract}

\section{Introduction}
\label{sec:intro}

\begin{figure}[htbp] 
  \centering 
  \includegraphics[width=0.38\textwidth]{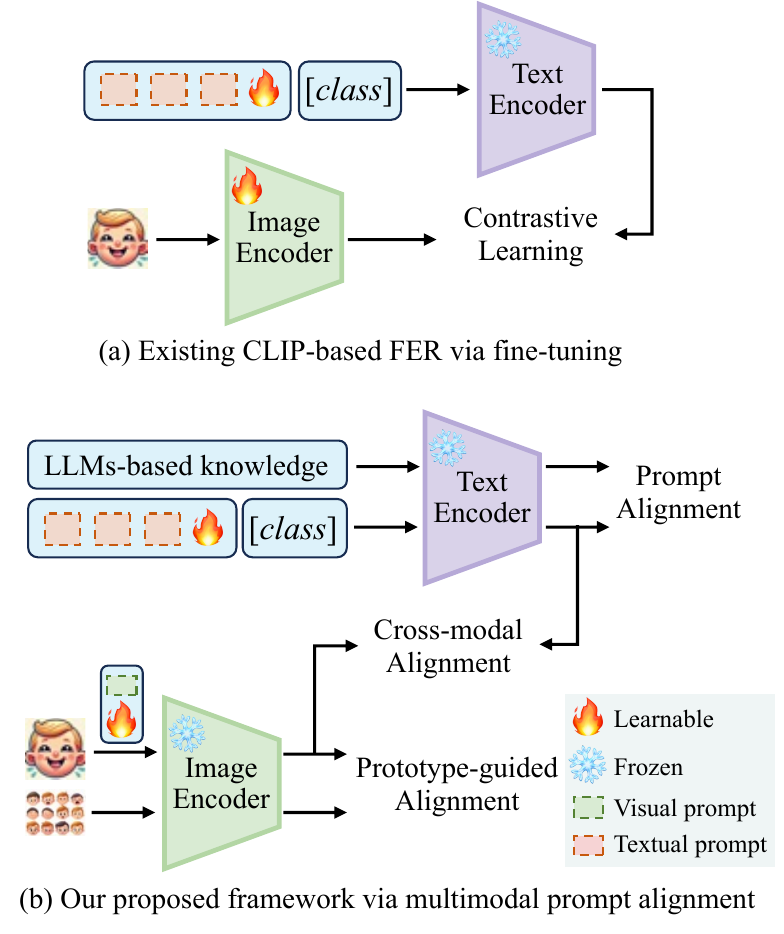} 
  \caption{(a) Conventional CLIP-based FER paradigm that adapts the pretrained model by fine-tuning the image encoder, which may lead to overfitting and diminished generalization.
  (b) Our proposed multimodal prompt alignment approach for FER, where the CLIP backbone remains frozen and learnable visual and textual prompts are used to achieve fine-grained cross-modal alignment, leveraging external language priors to enhance task-specific adaptation.
  }
  \label{figure1}
\end{figure}

Facial expression is one of the most common and natural ways for people to convey different emotions.
Understanding facial expressions has various emerging applications, such as human-robot interaction \cite{wimmer2008facial}, psychology study \cite{wallace2008investigation} and healthcare \cite{bisogni2022impact}.
Previous facial expression recognition (FER) methods have primarily been developed for highly controlled environments \cite{lucey2010extended, valstar2010induced, zhao2011facial} with consistent lighting, neutral backgrounds, and limited head poses, which may lack the robustness needed for real-world applications.
Despite the progress made by deep learning approaches \cite{zeng2018facial,li2020deep}, in-the-wild FER remains challenging due to the subtle and nuanced differences among facial expressions, varying lighting conditions, occlusions, and the presence of irrelevant background information.

Mainstream in-the-wild FER methods \cite{li2021adaptively,zhang2021weakly,li2017reliable,wang2020suppressing,wang2020region,ma2021facial,liu2024norface} rely on Convolutional Neural Networks (CNNs) \cite{lecun1989backpropagation} or Vision Transformers (ViTs) \cite{dosovitskiy2020image} to extract features from facial images, using one-hot labels for supervision during training. These methods have shown impressive performance, leveraging the power of deep networks to learn hierarchical spatial features from images.
However, a major limitation of these approaches is their reliance on one-hot labels, which treats each expression as a separate class without explicitly integrating the semantic relationships between these classes. 
As a result, the lack of semantic guidance hinders the model’s ability to generalize well, particularly when dealing with complex, real-world datasets where facial expressions might not always align with clear-cut class boundaries.

Advancements in vision-language models (VLMs) \cite{su2019vl, jia2021scaling, alayrac2022flamingo,kirillov2023segment}, particularly CLIP \cite{radford2021learning}, have demonstrated remarkable success in bridging the semantic gap between visual and textual modalities.
CLIP’s ability to learn powerful visual representations from large-scale data, coupled with its inherent generalization capabilities, has paved the way for adapting such models to specialized tasks like FER.
Recent works have explored prompt learning strategies \cite{khattak2023maple,wasim2023vita,zhou2022learning,zhao2023prompting} that adapt pretrained VLMs to downstream tasks (such as FER \cite{li2024cliper, zhou2024ceprompt, wang2023pose, li2024knowledge,li2024fer}) by learning task-specific textual prompts. 
However, these approaches still struggle to capture the fine-grained visual features necessary for accurate recognition.
As shown in \cref{figure1} (a), existing methods have the following limitations: integrating manually crafted text templates (e.g., “a~photo~of~[class]”) is often too coarse to capture the fine-grained visual cues that differentiate subtle facial expressions, and fine-tuning the entire image encoder risks overfitting and diminishing the generalization power of the pretrained model.

To overcome such limitations, we propose a novel multimodal prompt alignment framework for FER (MPA-FER) that enhances the recognition performance by integrating external knowledge and aligning multi-granularity visual and textual representations, as shown in \cref{figure1} (b).
In this work, our approach leverages large language models (LLMs), such as ChatGPT-3.5\cite{achiam2023gpt}, to generate detailed, class-specific descriptions that capture the critical visual patterns of facial expressions. 
These LLM-generated descriptions form multi-granularity hard textual prompts, which are then used to guide the learning of soft prompts. By aligning soft prompts with hard prompts at both token-level and prompt-level, our framework effectively incorporates external semantic knowledge into the adaptation process.

Furthermore, to preserve the generalization ability of the pretrained CLIP model while adapting it to the FER domain, we keep the CLIP model frozen and introduce a set of visual prompts to each frozen encoder layer, adding only negligible extra parameters. We employ class-specific prototypes, computed from the frozen visual features of the pretrained model, to steer the learning process and ensure that the prompted visual features remain aligned with the pretrained representation space.
These prototypes act as anchors that maintain the alignment of the visual features with the true semantics of each class, thereby eliminating the need to fine-tune the image encoder's core parameters and reducing the risk of overfitting.
Additionally, a cross-modal global-local alignment module is introduced to minimize the impact of class-irrelevant background regions. Our MPA-FER learns sparse, discriminative local features by aligning soft textual prompts with a sparse subset of facial regions, facilitating text-to-image matching that captures fine-grained semantics. By selectively focusing on these local features, we reduce noise from irrelevant regions and improve the model's ability to accurately distinguish subtle facial expressions, enhancing both its robustness and performance in the FER task.

Extensive experiments show that the proposed MPA-FER significantly boosts the FER performance by effectively integrating external semantic information and preserving the robust generalization capabilities of the pretrained CLIP model. In summary, the main contributions of this work are as follows:
\begin{itemize}
  \item We are the first to explore the learning paradigm of using a fully frozen CLIP model for FER, as opposed to the fine-tuning approach that has been widely used in VLM-based FER methods.
  \item We propose the multimodal prompt alignment framework for FER that effectively incorporates external semantic knowledge into the adaptation process by aligning soft prompts with multi-granularity hard textual prompts.
  \item We develop a prototype-guided visual feature alignment technique that regularizes the prompted visual features by anchoring them to class-specific prototypes derived from frozen CLIP features.
  \item Through extensive experiments on three in-the-wild FER datasets, we demonstrate that our framework not only outperforms state-of-the-art methods but also retains the generalization benefits of the pretrained CLIP model, while operating at minimal computational overhead.
\end{itemize}

\section{Related Work}
\label{sec:formatting}

\paragraph{Facial Expression Recognition}
Facial Expression Recognition (FER) has been an active research area for decades. Traditional FER methods largely rely on handcrafted features, such as Local Binary Patterns (LBP) \cite{shan2005robust} and Histogram of Oriented Gradients (HOG) \cite{dalal2005histograms}, to encode the geometry and texture changes in the face induced by different expressions.  However, these methods often fail to generalize well in unconstrained, in-the-wild settings where conditions such as lighting, occlusion, and varying head poses present significant challenges.
Additionally, these traditional methods struggle to capture the complex relationships between different facial features and the underlying emotions, limiting their performance on the FER task.
With the rise of deep learning, CNNs and ViTs have become the dominant approaches in FER, benefiting from their ability to learn hierarchical features from large amounts of data \cite{li2017reliable,li2021adaptively,xue2022vision,ma2023transformer}. 
For instance, Li \textit{et al.} \cite{li2018occlusion} used CNN-based attention mechanisms to focus on key regions of interest based on facial landmarks. Zhao \textit{et al.} \cite{zhao2021learning} developed a multi-scale attention network to capture both global and local features, improving robustness to occlusion and pose variations. Transformer-based methods, such as Xue \textit{et al.} \cite{xue2021transfer} and Ma \textit{et al.} \cite{ma2021facial}, apply multi-head self-attention to model relationships between different facial regions, enabling more flexible feature extraction. 
While these methods have shown impressive results, they rely exclusively on visual features and one-hot labels, which often suffer from issues like redundant feature learning, overfitting, and a lack of interpretability.

\paragraph{Vision-Language Models}
Vision-Language Models (VLMs), such as CLIP, have shown great success in leveraging large-scale multimodal data to learn joint representations of images and text.
For example, CoOp \cite{zhou2022learning} and its variants \cite{zhou2022conditional,miyai2023locoop,sun2022dualcoop} have used learnable prompt vectors to improve task-specific adaptations of VLMs. However, these methods often rely on generic textual prompts like “a~photo~of~[class]”, which lack the fine-grained detail necessary for distinguishing subtle facial expressions.
Several prior studies have also introduced external knowledge to prompt learning to enhance the quality of semantic guidance \cite{yao2023visual,bie2024xcoop,bulat2023lasp}. For instance, Yao \textit{et al.} \cite{yao2023visual} adopt the prompt template “a~photo~of~[class]” to guide the learning of soft prompts at the global level. While these approaches have shown promise, they often suffer from insufficient knowledge and inadequate guidance.
In the context of VLMs for FER \cite{zhang2023weakly, zhou2024ceprompt, li2024knowledge,li2024fer}, the focus is mainly on exploiting multimodal fusion techniques and textual prompts to improve facial expression recognition.
For example, Zhou \textit{et al.} \cite{zhou2024ceprompt} introduce expression class tokens with a Conception-Appearance Tuner (CAT) to create trainable appearance prompts, and enhance visual embeddings via knowledge distillation from CLIP.
Moreover, these VLM-based FER methods often adopt the fine-tuning paradigm for the pretrained CLIP image encoder to obtain better expression-related features and recognition performance.
However, this fine-tuning paradigm increases training costs and the risk of overfitting.
A more ideal approach should focus on improving the representation ability of the frozen backbone while enhancing cross-modal interaction as efficiently as possible. This principle underpins the approach we adopt in this work.

\section{Methodology}

\begin{figure*}[htbp] 
  \centering 
  \includegraphics[width=\textwidth]{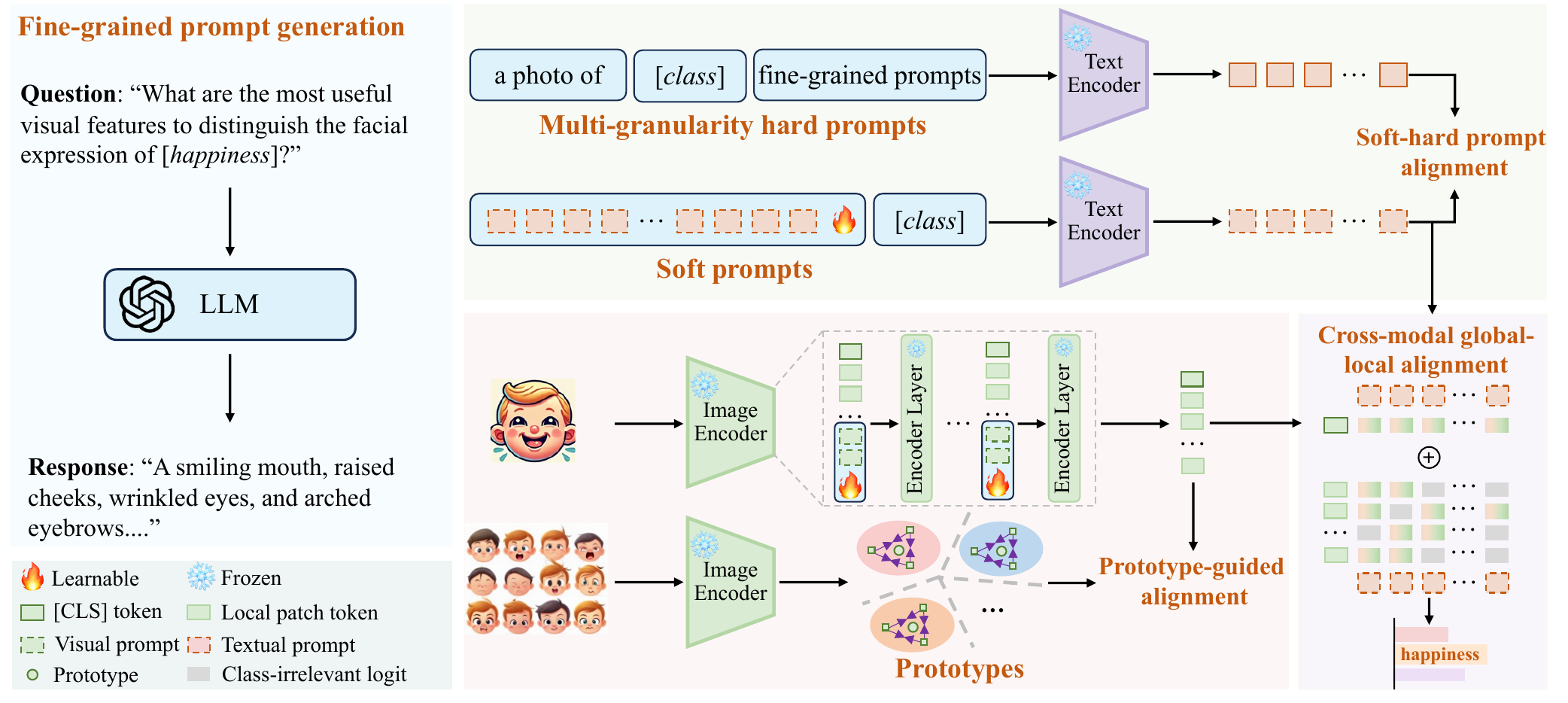} 
  \caption{An overview of our proposed multimodal prompt alignment framework, MPA-FER. The multi-granularity hard prompts consist of a generic template, category names, and LLM-generated class-specific descriptions. These hard prompts and frozen prototypes guide the learning of soft prompts and prompted visual features, respectively. The textual features of the soft prompts are aligned with the prompted visual features at both global and local levels, where sparse cross-modal similarities are computed to filter out class-irrelevant local features.
  }
  \label{method1}
\end{figure*}

\subsection{Overview}

In this work, we adopt the pretrained CLIP model as the vision-language foundation for FER, using a multimodal prompt alignment framework to preserve both strong generalization capability and high supervised performance. Our proposed method enables the use of an existing image-language pretrained model, eliminating the need to train one from scratch or fine-tune the entire encoders. An overview of the proposed MPA-FER is shown in \cref{method1}.
Our framework effectively integrates external semantic knowledge and aligns multi-granularity textual and visual representations to enhance FER performance.

At the core of our method, we use the prompt learning scheme on both the text encoder and the image encoder.
On the text side, we model the context words using trainable soft prompts, as opposed to handcrafting textual inputs based on class labels.
To guide the learning of soft prompts, we use multi-granularity hard prompts, which consist of a generic template, class category names, and class-specific descriptions generated by a LLM.
These detailed descriptions capture the unique visual features of each facial expression and provide task-specific semantic guidance.

On the vision side, we introduce a set of visual prompts into the output feature sequence from each encoder layer, learning face-specific context vectors.
To ensure that the prompted visual features remain aligned with the pretrained representation space, we utilize frozen prototypes derived from the pretrained CLIP model. These class-specific prototypes act as anchors, ensuring that the prompted visual features keep consistent with the true semantics of each class. 
Additionally, our framework incorporates a cross-modal global-local alignment module to align the representations of soft textual prompts with the prompted visual features at both global and local levels. At the local level, we compute sparse cross-modal similarities to focus on the most relevant facial regions (such as the eyes, mouth, and brows), while filtering out irrelevant features from the background.

\subsection{Fine-grained Prompt Generation}

Although existing prompt-tuning schemes \cite{zhou2022learning, zhou2022conditional, li2024cliper} can effectively adapt the pretrained CLIP model to downstream tasks, they often fail to capture the fine-grained visual features across different facial expressions, leading to underperformance compared to state-of-the-art FER methods. To address this limitation, we propose leveraging external knowledge as language priors and enhancing soft prompting by integrating multi-granularity hard prompts.
The pipeline of our framework is illustrated in \cref{method1}. To explicitly incorporate external knowledge into the soft prompting process, we use a large language model (LLM), such as ChatGPT-3.5, to generate detailed and comprehensive descriptions for each facial expression category. Specifically, we design a query to prompt the LLM: “What are the most useful visual features to distinguish the facial expression of $[class]$?". In response, the LLM generates descriptions in the format: “The most useful visual features for distinguishing the facial expression of $[class]$ include: \dots".
We then construct class-specific, multi-granularity hard prompts by combining a generic text template with the facial expression name and the corresponding LLM-generated descriptions. This multi-granularity design ensures that both coarse-grained and fine-grained semantics are effectively integrated, further enhancing the adaptability and performance of our proposed framework.

\subsection{Soft-hard Prompt Alignment}

Previous prompt-tuning methods primarily focus on aligning visual-textual semantics and capturing cross-modal interactions via contrastive learning. In contrast, we additionally model text-text interactions and enhance soft prompt learning by integrating multi-granularity hard prompts through a soft-hard prompt alignment mechanism. Since the hard prompts contain fine-grained facial expression descriptions, this alignment explicitly injects external knowledge into the soft prompts by minimizing the discrepancy between the soft and hard prompts.

Soft prompts consist of a small set of learnable parameters that are prone to overfitting on downstream tasks \cite{bulat2023lasp,park2024prompt}. 
To address this, we leverage the LLM-based hard prompts to regularize the soft prompts, encouraging them to remain close in the embedding space to the hard prompts.
Specifically, for each class, the generated hard prompts are passed through the tokenizer to get token embeddings for each word in the hard prompt.
Given the token embeddings of the soft prompts and the multi-granularity hard prompts, $\bm{t}_c$ and $\bm{t}_c^*$, we first align them at the token level via contrastive learning.
Under the guidance of the hard prompts, the soft prompts are trained so that they can be correctly classified in the embedding space according to the class weights provided by the hard prompts.
Consequently, the probability distribution of the soft prompt embeddings $\bm{t}_c$ over the class labels is defined as
\begin{equation}
  P(y_d|\bm{t}_d) = \frac{\exp (sim(\bm{t}_d,\bm{t}_d^*)/\tau)}{\sum_{c=1}^{C}\exp (sim(\bm{t}_c,\bm{t}_d^*)/\tau)},
\end{equation}
where $y_d$ denotes the binary label of class $d$,  \( sim(\cdot) \) represents the cosine similarity, $C$ denotes the number of facial expressions, and \( \tau \) is a temperature parameter.
Finally, the token-level alignment loss $\mathcal{L}_{ta}$ can be optimized by minimizing the cross entropy loss:
\begin{equation}
  \mathcal{L}_{ta} =  -\sum_{c=1}^{C}y_c\log P(y_c|\bm{t}_d).
  \label{loss1}
\end{equation} 

Furthermore, the discrepancy between learnable knowledge and LLM-based knowledge can be measured by the distance between the corresponding textual embeddings.
The textual embeddings of the hard prompts and the soft prompts generated by the text encoder, $\theta$, can be denoted as $\theta(\bm{t}_k^*)$ and $\theta(\bm{t}_k)$, respectively.
To apply the soft-hard prompt alignment at the prompt level, we formulate it as follows:
\begin{equation}
\begin{aligned}
  P(y_d|\theta(\bm{t}_d)) &= \frac{\exp (sim(\theta(\bm{t}_d),\theta(\bm{t}_d^*))/\tau)}{\sum_{c=1}^{C}\exp (sim(\theta(\bm{t}_c),\theta(\bm{t}_d^*))/\tau)},\\
  \mathcal{L}_{pa} &=  -\sum_{c=1}^{C}y_c\log P(y_c|\theta(\bm{t}_d)),
\end{aligned}
\end{equation}
where $\mathcal{L}_{pa}$ represents the prompt-level alignment loss. Thus, the soft-hard textual prompt alignment loss $\mathcal{L}_{t}$ can be computed as the sum of $\mathcal{L}_{pa}$ and $\mathcal{L}_{ta}$:
\begin{equation}
  \mathcal{L}_{t} =  \mathcal{L}_{ta} + \mathcal{L}_{pa}
  \label{loss_t}
\end{equation}
\cref{loss_t} ensures that both token-level and prompt-level textual alignment are effectively optimized.

\subsection{Prototype-guided Visual Feature Alignment}

To better retain the general representation of the pretrained image encoder, we introduce the prompt learning scheme into the image encoder that keeps the pretrained model frozen.
For prompting on the image encoder, we have two major objectives: 1) Providing additional parameters to adapt the CLIP image representations towards the FER dataset distribution, and 2) ensuring the prompted visual features align well with the pretrained CLIP features to address the prompt overfitting problem for better generalization.

In this regard, we introduce a set of visual prompts, which requires minimal additional parameters to train, into the image encoder $\phi$ that learns face-specific context vectors.
The prompt tokens are randomly initialized learnable vectors, and they are designed to provide additional learning capacity to adapt the pretrained model to the FER dataset distribution.
Specifically, the prompt tokens $\{p_{i}^{l}\}_{i=1}^{N_p}$ at $l$-th layer are appended to the visual token sequence $\bm{z}^{l-1}$ before applying the frozen pretrained encoder layer:
\begin{equation}
  [\bm{z}^{l},\_] =  \phi_{l}(\left[ \bm{z}^{l-1}, \{p_{i}^{l}\}_{i=1}^{N_p} \right]), l=1,2,3,\dots,K
\end{equation}
where ${N_p}$ and $K$ are the number of visual prompts and the depth of the image encoder $\phi$, respectively.
During training, the visual prompts act as contextual modifiers to interact with these pretrained visual feature tokens.
Finally, the prompted visual features $\bm{z}^{K}$ are utilized to learn a common feature space with textual labels.

To retain the generalized knowledge of CLIP features and further optimize the visual prompts more explicitly, we impose the regularization constraint on the prompted visual features and the CLIP visual prototypes.
For the pretrained CLIP visual encoder, we define the prototypes $\bm{p}$ as the mean of the frozen feature representations from the subset of training dataset $\mathcal{D}_{train}=\{{ (x_i,y_i )}\}_{i=1}^{N_{train}}$ for each class $c$:
\begin{equation}
  \bm{p}_c = \frac{1}{N^{c}_{subset}}\sum_{(x_i,y_i)\in \mathcal{D}_{subset}^{c}}\bm{z}^{g}_{i},
\end{equation}
where \( \mathcal{D}_{subset} \) denotes a subset of \( \mathcal{D}_{train} \), \( N^{c}_{subset} \) represents the number of samples belonging to class \( c \) in \( \mathcal{D}_{subset} \). $\bm{z}^{g}$ denotes the global classification token from the output sequence of the final image encoder layer. The channel axis in $\bm{z}^g$ is omitted for simplicity.

Therefore, we can explicitly condition the prompted visual features to be close to their corresponding prototype as follows,
\begin{equation}
  \mathcal{L}_{v} = \sum_{c=1}^{C} \sum_{i}^{N_{train}} \mathbb{I}_{(y_i=c)} \mathcal{M}(\bm{z}^{g},\bm{p}_c),
  \label{loss_v}
\end{equation}
where $\mathcal{M}$ denotes the matching losses (such as L1 loss and cosine similarity) to measure the distance between the prompted classification token and its corresponding prototype.
Since we do not introduce any trainable parameters for generating the frozen prototypes, the prototype-guided visual feature alignment not only regularizes the visual prompts to enhance generalization but also operates with minimal computational overhead.

\subsection{Cross-modal Global-local Alignment}
Different facial expressions are reflected in specific regions of the face, such as the eyes, mouth, and brow, while other areas, like the background, contribute little to the expression.
To better model the relationship between textual representations and visual features, it is essential to focus on these expression-relevant local regions, rather than simply averaging similarities across all spatial locations. 
To this end, we propose a cross-modal global-local alignment module to align soft textual prompts and prompted visual features.
Specifically, the global visual representation $\bm{z}^{g}$ and a set of local visual features $\bm{Z}^{l}=\{ \bm{z}^{l}_{i} \}_{i=1}^{N_l}$ are used to align with the textual representation $\theta(\bm{t}_d)$.
We further utilize the sparse local similarity to learn sparse discriminative local features, where only local features semantically related to the class are kept to perform classification:
\begin{equation}
  sim_{top-k}(\bm{Z}^{l},\theta(\bm{t}_d)) = \frac{1}{k}\sum_{i=1}^{N_l}\mathbb{I}_{top-k}(i)\cdot\langle \bm{z}^{l}_{i},\theta(\bm{t}_d)\rangle,
\end{equation}
\begin{equation}
  \mathbb{I}_{top-k}(i) = \begin{cases}
    1, & \text{if rank}_{i} (\langle \bm{z}^{l}_{i},\theta(\bm{t}_d)\rangle) \le k, \\
    0, & \text{otherwise},
    \end{cases}
\label{topk}
\end{equation}
where $\mathbb{I}_{top-k}(l)$ denotes the indicator function for selecting the top-k local features with the highest similarities with the textual representation, and ${N_l}$ is the number of local features.
This selective focus reduces noise from irrelevant regions and ensures that the learning process emphasizes the facial features that are most indicative of the expression, thereby improving the accuracy and robustness of the model.
The combination of global similarities and sparse local similarities produces the output logits:
\begin{equation}
  {logits} = sim(\bm{z}^{g}, \theta(\bm{t}_d)))+sim_{top-k}(\bm{Z}^{l},\theta(\bm{t}_d)) \label{loss4}.
\end{equation}
Thus, the cross-modal alignment can be optimized using cross-entropy loss by minimizing the discrepancy between the predicted results and the ground truth:
\begin{equation}
 \mathcal{L}_{v\_t}= CE(logits,y_d),
\end{equation}
where $\mathcal{L}_{v\_t}$ denotes the image-text alignment loss, and $CE$ is the stand cross entropy loss.

Combining the textual prompt alignment loss in \cref{loss_t}, the visual feature alignment loss in \cref{loss_v}, the cross-modal alignment loss in \cref{loss4}, our method optimize the text prompts and the visual prompts by minimizing the following total loss:
\begin{equation}
  \mathcal{L}_{total} = \mathcal{L}_{v\_t} + \beta \times \mathcal{L}_{t} + \gamma \times \mathcal{L}_{v},
\label{totalloss}
\end{equation}
where $\beta$, $\gamma$ are hyper-parameters to balance the losses.

\begin{table}[t]
  \centering
  \resizebox{1.0\linewidth}{!}{
  \begin{tabular}{l|cc}
  \toprule
  \textbf{Method} & \textbf{RAF-DB}      & \textbf{AffectNet-7} \\
  \midrule
  Baseline CoOp \cite{zhou2022learning} & 86.15 & 62.27   \\
  +~Visual Prompts  & 88.47 \textcolor{ForestGreen}{(+2.32)}& 63.90 \textcolor{ForestGreen}{(+1.63)} \\
  ~~~~+~Prototype-guided Align.  & 89.42 \textcolor{ForestGreen}{(+0.95)} & 65.02 \textcolor{ForestGreen}{(+1.12)} \\
  ~~~~~~~~+~Soft-hard Prompt Align. & 91.18 \textcolor{ForestGreen}{(+1.76)} & 66.58 \textcolor{ForestGreen}{(+1.56)}\\
  ~~~~~~~~~~~~+~Cross-modal Align. & \textbf{92.51} \textcolor{ForestGreen}{(+1.33)} & \textbf{67.85} \textcolor{ForestGreen}{(+1.27)} \\
  \bottomrule
  \end{tabular}%
  }
  \caption{Ablation study on the contribution of individual components in our MPA-FER framework for FER.
  }
  \label{tab:ablation_ours}%
  \end{table}%

\section{Experiments Settings}
\paragraph{Datasets}
i) \textbf{RAF-DB} \cite{li2017reliable}: consists of 29,672 real-world facial images collected from the internet. Each image is annotated by 40 annotators with basic expressions or compound emotions. For our experiments, we focus on the common expression labels, which include 12,271 training samples and 3,068 testing samples.
ii) \textbf{FERPlus} \cite{barsoum2016training}: is an extended version of FER2013, containing 28,709 training images, 3,589 validation images, and 3,589 testing images. Each image in FERPlus is annotated by 10 workers across eight expression categories. For fairness, we use majority voting to generate the final labels.
iii) \textbf{AffectNet} \cite{mollahosseini2017affectnet}: contains over one million facial images gathered from the internet. Previous studies have focused on AffectNet with either 7 or 8 facial expression categories. We refer to these versions as AffectNet-7 and AffectNet-8, respectively. For our experiments, we use approximately 280,000 images for training, along with 3,500 validation images for AffectNet-7 and 4,000 validation images for AffectNet-8.

\paragraph{Implementation Details}
In our experiments, both the text and image encoders remain frozen, with their weights directly inherited from the pretrained CLIP models (ViT-B/16 or ViT-L/14).
The visual prompts and soft textual prompts are randomly initialized from a Gaussian distribution with a zero mean and 0.02 std.
The visual prompt number $N_p$ and the textual prompt number are set to 8 and 10, respectively.
The training epochs are 120 for RAF-DB and FERPlus, 20 for AffectNet with a batch size of 32. We use SGD optimizer with a learning rate of 0.032 and adopt a cosine decay strategy for the learning rate.
During the training process, all images are resized to $224\times224$ pixels, and several standard data augmentation techniques, such as random cropping and erasing, are employed on the training images.
The hyper-parameter $k$ used for local features selection in \cref{topk} is set to 16.
In \cref{totalloss}, we experimentally choose suitable loss balancing factors as $\beta = \gamma = 1$.
We implement our method with the PyTorch framework and train the models on one NVIDIA V100 GPU with 32GB RAM.

\subsection{Ablation Study}

\begin{table}[t]
  \centering
  \resizebox{1.0\linewidth}{!}{
  \begin{tabular}{l|cc}
  \toprule
  \textbf{Textual Prompt Configuration} & \textbf{RAF-DB} & \textbf{FERPlus} \\
  \midrule
  CoOp + (1) & 86.15 & 87.48 \\
  CoOp + (2) & 86.59 & 88.01 \\
  CoOp + (3) & 87.26 & 88.30 \\
  CoOp + (1) + Soft-hard Prompt Align. & 87.56 & 88.88 \\
  CoOp + (2) + Soft-hard Prompt Align.& 87.81 & 89.21 \\
  CoOp + (3) + Soft-hard Prompt Align. & 88.69 & 89.76 \\
  \midrule
  MPA-FER + (1) & 89.60 & 88.77 \\
  MPA-FER + (2) & 90.48 & 88.94 \\
  MPA-FER + (3) & 91.01 & 89.69 \\
  MPA-FER + (1) + Soft-hard Prompt Align. & 91.05 & 90.17 \\
  MPA-FER + (2) + Soft-hard Prompt Align. & 91.87 & 90.43 \\
  MPA-FER + (3) + Soft-hard Prompt Align. & \textbf{92.51} & \textbf{91.15} \\
  \bottomrule
  \end{tabular}
  }
  \caption{Ablation study on different textual prompt configurations and the effect of soft-hard prompt alignment.
  }
  \label{tab:prompt_ablation}
\end{table}

\paragraph{Effect of Different Model Components.}
In our ablation study (see \cref{tab:ablation_ours}), we systematically evaluate the impact of each key component in our MPA-FER framework for FER. Starting with a baseline model that employs a frozen CLIP image encoder trained solely with contrastive loss, we first introduce the visual prompts, which improved accuracy by 2.32\% on RAF-DB and 1.63\% on AffectNet-7. Next, adding the prototype-guided visual alignment module yields further gains of 0.95\% and 1.12\%, respectively. The subsequent integration of soft-hard prompt alignment results in additional improvements of 1.76\% on RAF-DB and 1.56\% on AffectNet-7, while the final inclusion of the cross-modal alignment module achieves the best overall performance with further enhancements of 1.33\% and 1.27\%. These results demonstrate that each component contributes meaningfully to performance, validating the effectiveness of our multimodal prompt alignment strategy for FER.

\begin{figure}[t] 
  \centering 
  \includegraphics[width=0.48\textwidth]{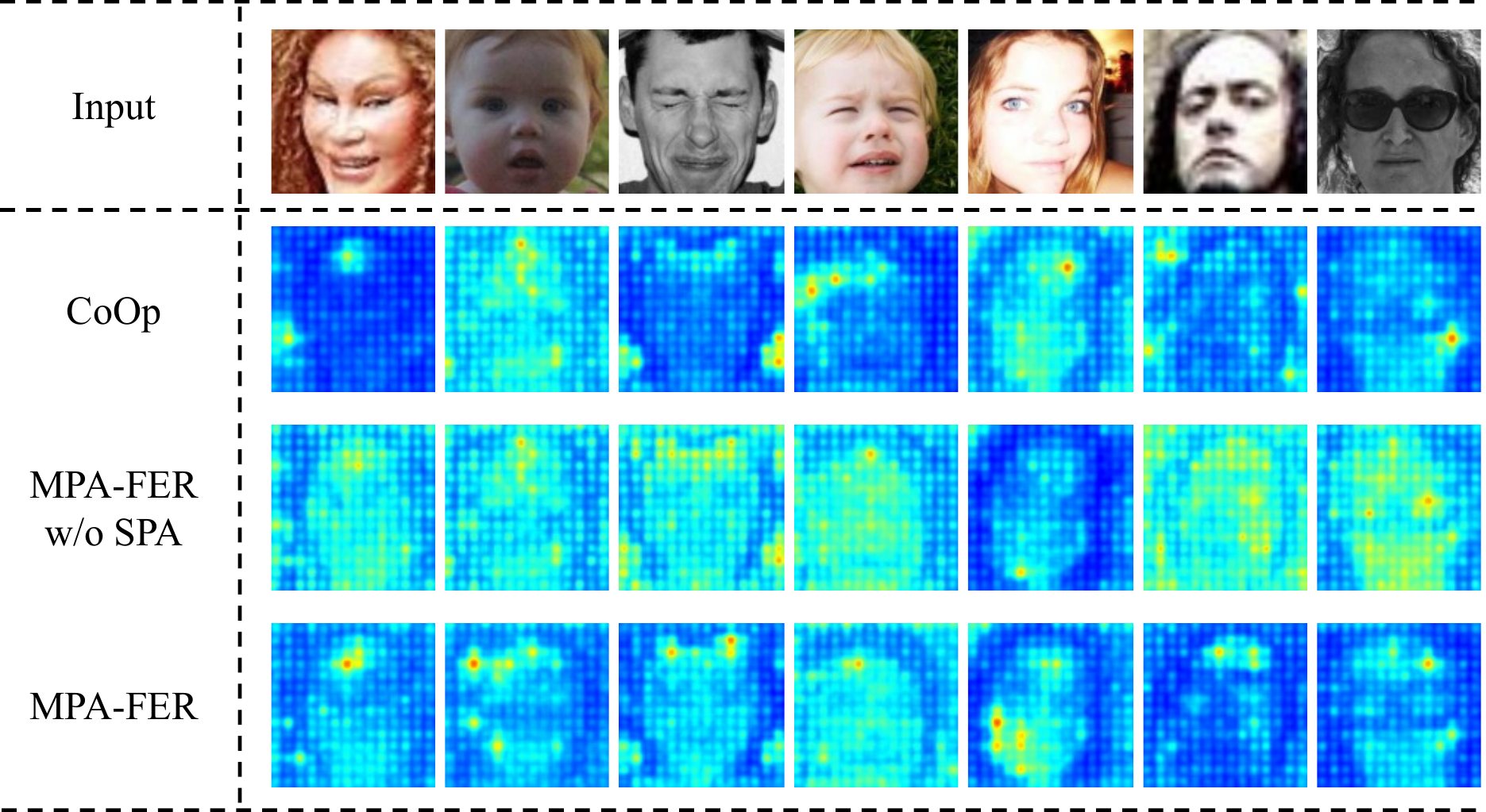} 
  \caption{The attention visualization of CoOp, MPA-FER without soft-hard prompt alignment (SPA) and our MPA-FER. We obtain these visualizations by backpropagating gradients on the input images. The first row displays the input images, while the second row shows the attention maps from CoOp, which only fine-tunes language prompts but is relatively sensitive to background features. The third row illustrates the attention maps from our MPA-FER without SPA, where the introduction of visual prompts begins to shift the focus towards expression-relevant features. The last row presents the results of our MPA-FER, which further emphasizes discriminative local facial regions and significantly reduces the influence of irrelevant background areas.
  }
  \label{figure3}
\end{figure}

\paragraph{Effect of Different Textual Prompts.} We further evaluate the impact of various textual prompt configurations on the performance of our proposed MPA-FER framework. In our experiments, we consider three types of hard prompts: (1) the conventional prompt “a~photo~of~[class]”, (2) a more semantically enriched prompt “a~photo~of~a~person~making~a~facial~expression~of~[class]”, and (3) an augmented prompt “a~photo~of~a~person~making~a~facial~expression~of~[class],~[LLM-based~descriptions]”, where the LLM-based descriptions provide detailed, class-specific visual cues. As shown in \cref{tab:prompt_ablation}, both the baseline CoOp and our MPA-FER framework benefit from improved prompt configurations, with performance increasing from configuration (1) to (3) on both RAF-DB and FERPlus.
Furthermore, incorporating soft-hard prompt alignment consistently enhances accuracy across all configurations. For example, our MPA-FER framework with configuration (3) and soft-hard prompt alignment achieves the highest accuracy of 92.51\% on RAF-DB and 91.15\% on FERPlus.
These results demonstrate that explicitly incorporating the hard prompts with class-specific descriptions significantly improves semantic guidance, thereby enhancing overall performance.
We also visualize the attention maps produced by different models (i.e., the baseline CoOp, our MPA-FER without SPA, and the MPA-FER) in \cref{figure3} for an intuitive visual comparison.

\begin{figure*}[htbp]
  \centering
  \begin{minipage}{0.33\textwidth}
    \centering
    \includegraphics[width=\textwidth]{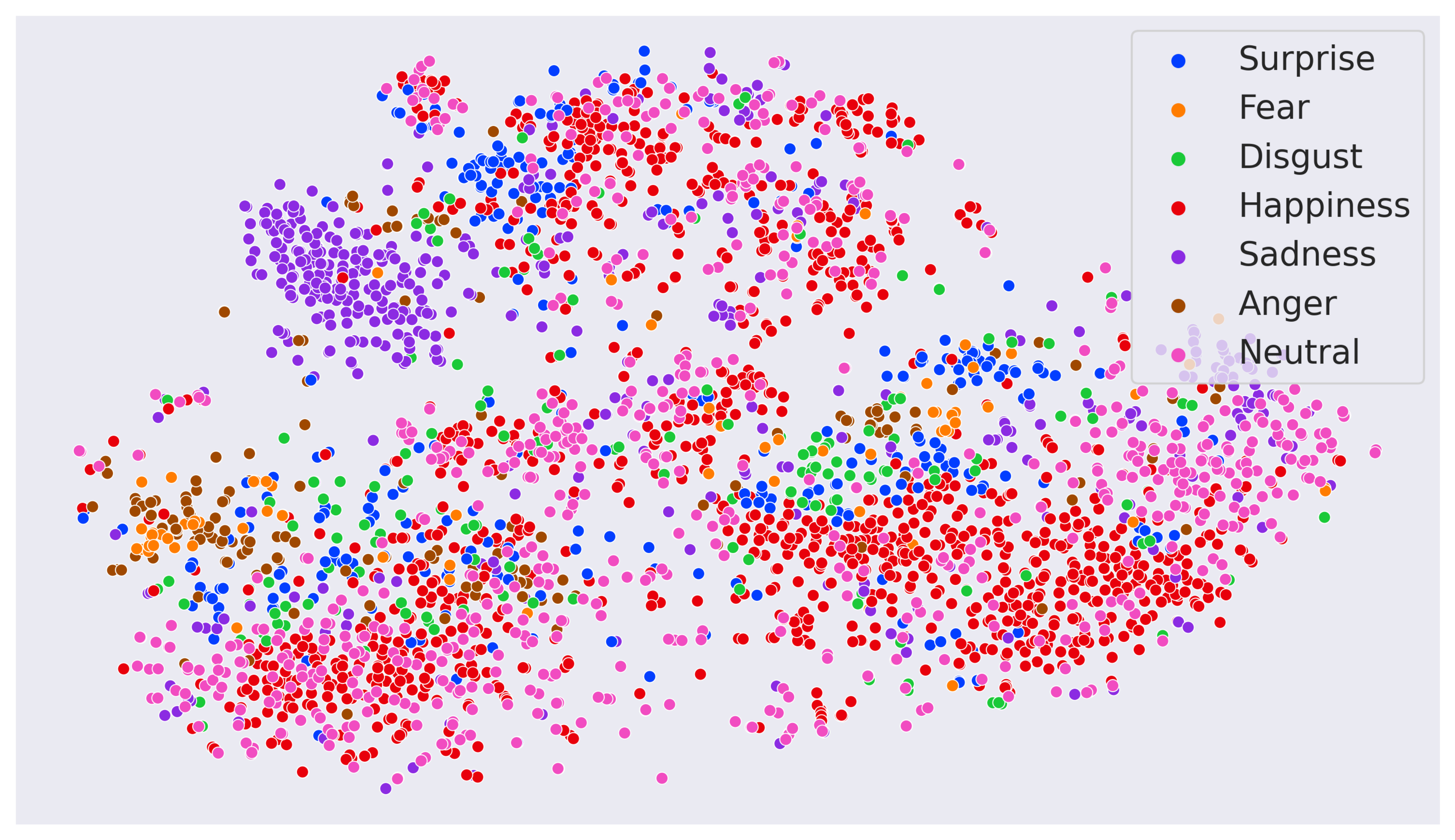}
    \subcaption{MPA-FER w/o Visual Prompting}
    \label{fig:image1}
  \end{minipage}%
  \begin{minipage}{0.33\textwidth}
    \centering
    \includegraphics[width=\textwidth]{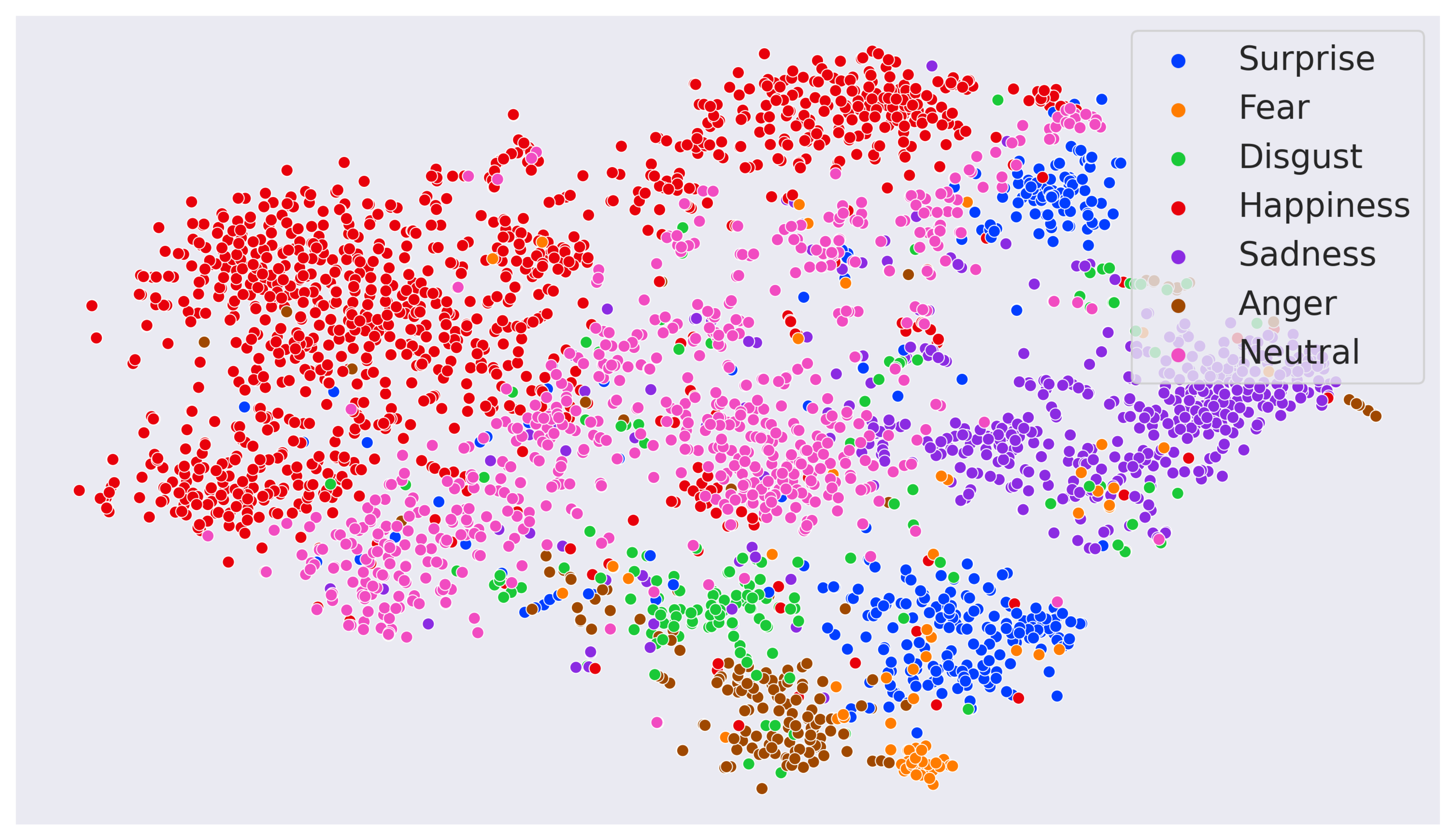}
    \subcaption{MPA-FER w/o Prototype-guided Alignment}
    \label{fig:image2}
  \end{minipage}%
  \begin{minipage}{0.33\textwidth}
    \centering
    \includegraphics[width=\textwidth]{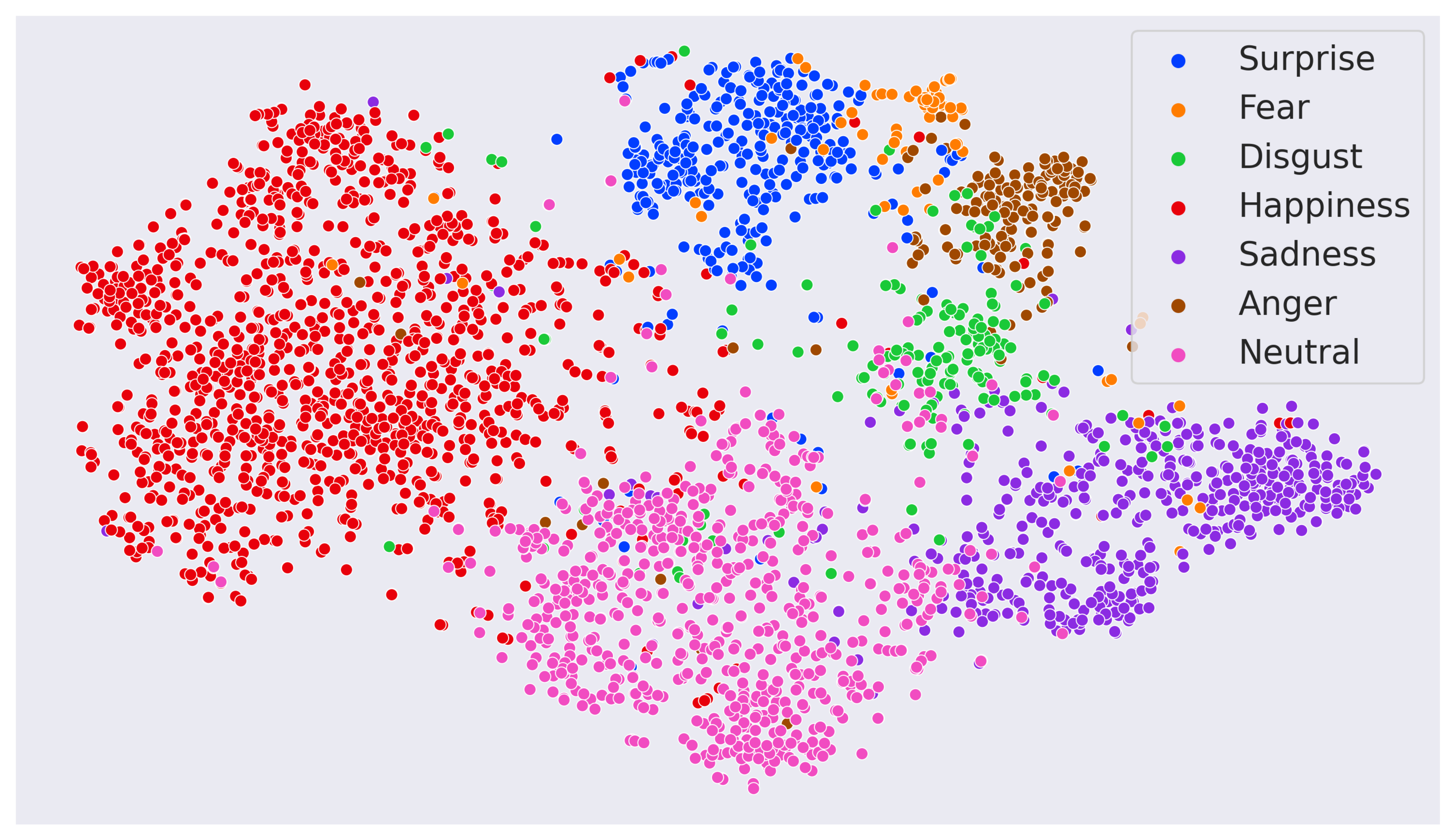}
    \subcaption{Our MPA-FER}
    \label{fig:image3}
  \end{minipage}
  \caption{t-SNE visualizations of the learned feature representations of the test set of RAF-DB. Subfigure (a) illustrates the results when the CLIP image encoder is used without any visual prompts, while subfigure (b) shows the feature distribution when the prototype-guided alignment module is omitted. Subfigure (c) presents the embeddings from our complete MPA-FER.}
  \label{fig:three_images}
\end{figure*}

\paragraph{Effect of Varying Training Data for Class-wise Prototypes.}
We conduct an ablation study by varying the number of training images per class used for generating class-wise prototypes from the training dataset. The experimental results in \cref{tab:prototype} demonstrate that increasing the number of training images per class for generating class-wise prototypes consistently improves performance on both datasets. For example, on the FERPlus dataset, accuracy increases from 89.87\% with a single image per class to 91.15\% when using the full training set.
These results indicate that using more training samples helps to capture more robust and representative class-specific features, although the improvements tend to plateau as the number of samples approaches the full dataset. In \cref{fig:three_images}, we visualize the t-SNE embeddings of the learned feature representations from three variants of our proposed framework. As can be observed, the full model yields more distinct and tightly clustered features, highlighting the critical contributions of both components in enhancing the discriminative power of the learned representations for FER.

\begin{table}[t]
  \centering
  \resizebox{1.0\linewidth}{!}{
  \begin{tabular}{l|cccccc}
  \toprule
  \textbf{Number} & \textbf{1} & \textbf{4} & \textbf{16} & \textbf{32} & \textbf{64} & \textbf{Full} \\
  \midrule
  FERPlus & 89.87 & 90.20 & 90.48 & 90.75 & 91.11 & \textbf{91.15} \\
  AffectNet-8 & 61.41 & 61.68 & 62.10 & 62.59 & 62.77 & \textbf{62.80} \\
  \bottomrule
  \end{tabular}
  }
  \caption{Ablation study on the varying amounts of training images per class for generating class-wise prototypes. (“\textbf{Full}” indicates the utilization of the entire training dataset.)
  }
  \label{tab:prototype}
\end{table}

\begin{table}[t]
  \centering
  \resizebox{1.0\linewidth}{!}{
  \begin{tabular}{l|ccccc}
  \toprule
  \textbf{Backbone} & \textbf{Parameters (MB)}  & \textbf{RAF-DB} & \textbf{AffectNet-7} & \textbf{AffectNet-8} & \textbf{FERPlus} \\
  \midrule
  ViT-B/16   & 0.218  & 92.51 & 67.85 & 62.80 & 91.15 \\
  ViT-L/14 & 0.443  & 93.74 & 68.89 & 63.74 & 91.81 \\
  \bottomrule
  \end{tabular}
  }
  \caption{Effectiveness of our proposed method with different CLIP backbone networks. We only count the number of parameters that require gradient updates.
  }
  \label{tab:complexity}
\end{table}

\paragraph{Effect of Working with Different CLIP Backbones.}
\cref{tab:complexity} shows the effectiveness of our proposed method using two different CLIP backbone networks: ViT-B/16 and ViT-L/14. With ViT-B/16, our model achieves 92.51\% on RAF-DB, 67.85\% on AffectNet-7, 62.80\% on AffectNet-8, and 91.15\% on FERPlus, with a total of 0.218 MB learnable parameters. When using the larger ViT-L/14 backbone, which requires only a modest increase in learnable parameters (0.443 MB), the performance improves consistently across all datasets (93.74\% on RAF-DB, 68.89\% on AffectNet-7, 63.74\% on AffectNet-8, and 91.81\% on FERPlus). These results demonstrate that our method scales well with larger CLIP backbones, yielding enhanced accuracy while maintaining a minimal increase in model complexity.

\begin{table}[t]
  \centering
  \resizebox{0.99\linewidth}{!}{
  \begin{tabular}{l|c|cc}
  \toprule
  \textbf{Method} & \textbf{Publication}  & \textbf{RAF-DB} & \textbf{FERPlus} \\
  \midrule
  \textit{CNN-based FER} \\
  RAN\cite{wang2020region}   & TIP 2020  & 86.90 & 88.55  \\
  KTN\cite{li2021adaptively} & TIP 2021  & 88.07 & 90.49  \\
  MA-Net\cite{zhao2021learning} & TIP 2021  & 88.40 & -  \\
  \midrule
  \textit{Transformer-based FER} \\
  VTFF\cite{ma2021facial} & TAFFC 2021  & 88.14 & 88.81  \\
  TransFER\cite{xue2021transfer} & ICCV 2021  & 90.91 & 90.83  \\
  APViT\cite{xue2022vision} & TAFFC 2022  & 91.98 & 90.86  \\
  TAN\cite{ma2023transformer} & TAFFC 2023  & 89.12 & 90.67  \\
  \midrule
  \textit{CLIP-based FER} \\
  CLEF \cite{zhang2023weakly} & ICCV2023  & 90.09 & 89.74  \\
  CLIPER \cite{li2024cliper} & ICME 2024  & 91.61 & -  \\
  FER-former\cite{li2024fer} & TMM 2024  & 91.30 & 90.96  \\
  E2NT\cite{li2024knowledge} & arXiv 2024  & 92.63 & 91.18  \\
  CEPrompt \cite{zhou2024ceprompt} & TCSVT 2024  & 92.43 & - \\
  Our MPA-FER & - & \textbf{93.74} & \textbf{91.81}\\
  \bottomrule
  \end{tabular}
  }
  \caption{Comparison with state-of-the-art methods on RAF-DB and FERPlus.
  }
  \label{tab:rafdb-ferplus}
\end{table}

\subsection{Comparison with the State-of-the-Arts}

\cref{tab:rafdb-ferplus} and \cref{tab:affectnet} present comprehensive comparisons of our proposed MPA-FER with state-of-the-art methods across multiple FER datasets. In \cref{tab:rafdb-ferplus}, our method achieves 93.74\% on RAF-DB and 91.81\% on FERPlus, outperforming traditional CNN-based approaches (e.g., RAN, KTN), recent Transformer-based methods (e.g., VTFF, TransFER, APViT, TAN), as well as existing CLIP-based methods (e.g., CLEF, CLIPER, FER-former, E2NT, CEPrompt). These results underscore the effectiveness of our multimodal prompt alignment framework, which leverages external semantic knowledge to enhance visual-textual representation without fine-tuning the pretrained CLIP model.

Similarly, \cref{tab:affectnet} demonstrates that our MPA-FER consistently outperforms other methods on AffectNet-7 and AffectNet-8. Our approach achieves 68.89\% on AffectNet-7 and 63.74\% on AffectNet-8, thereby confirming its robustness and generalizability under challenging, in-the-wild conditions. Overall, these comparisons validate that MPA-FER not only improves performance over traditional and recent deep learning-based FER methods but also sets a new state-of-the-art in the field by efficiently leveraging multimodal information.
See more experiments in the Supplementary Material.

\begin{table}[t]
  \centering
  \resizebox{0.99\linewidth}{!}{
  \begin{tabular}{l|c|cc}
  \toprule
  \textbf{Method} & \textbf{Publication}  & \textbf{AffectNet-7} & \textbf{AffectNet-8} \\
  \midrule
  SCN \cite{wang2020suppressing} & CVPR 2020  & 63.40 & 60.23  \\
  MA-Net\cite{zhao2021learning} & TIP 2021  & 64.53 & 60.29  \\
  VTFF\cite{ma2021facial} & TAFFC 2021  & 64.80 & 61.85  \\
  MVT \cite{li2021mvit} & arXiv 2021  & 64.57 & 61.40  \\
  POSTER\cite{zheng2023poster} & ICCV 2023  & 67.31 & 63.34  \\
  CLEF \cite{zhang2023weakly} & ICCV2023  & 65.66 & 62.77  \\
  CLIPER \cite{li2024cliper} & ICME 2024  & 66.29 & 61.98  \\
  CEPrompt \cite{zhou2024ceprompt} & TCSVT 2024  & 67.29 & 62.74 \\
  Our MPA-FER & - & \textbf{68.89} & \textbf{63.74}\\
  \bottomrule
  \end{tabular}
  }
  \caption{Comparison with state-of-the-art methods on AffectNet-7 and AffectNet-8.
  }
  \label{tab:affectnet}
\end{table}

\section{Conclusion}
In this paper, we introduce the multimodal prompt alignment framework for facial expression recognition (MPA-FER) that leverages external semantic knowledge without requiring extensive fine-tuning of the pretrained CLIP model. By incorporating LLM-based hard prompts and aligning them with trainable soft prompts, our approach captures fine-grained facial cues to enhance visual-textual representation learning. Additionally, our prototype-guided feature alignment and cross-modal alignment modules ensure that the prompted visual features remain robust and discriminative by focusing on the most expression-relevant regions while suppressing irrelevant background information.
Extensive experiments on in-the-wild FER datasets demonstrate that MPA-FER outperforms state-of-the-art methods, achieving superior accuracy with adding few learnable parameters. The ablation studies further validate the contributions of each component within our framework, highlighting its effectiveness and efficiency.
Future work will explore further integration of multimodal cues and the extension of our framework to other fine-grained visual recognition tasks.

{
    \small
    \bibliographystyle{ieeenat_fullname}
    \bibliography{main}
}

\end{document}